\RequirePackage{iftex}\ifpdftex\pdfoutput=1\fi
\documentclass[11pt]{article}
\usepackage[a4paper,margin=1in]{geometry}
\usepackage[T1]{fontenc}
\usepackage{lmodern}
\usepackage{microtype}
\usepackage{graphicx}
\usepackage{booktabs}
\usepackage{array}
\usepackage{xcolor}
\usepackage{listings}
\usepackage{float}
\usepackage{caption}
\usepackage{subcaption}
\usepackage{tikz}
\usetikzlibrary{arrows.meta,positioning,fit}
\usepackage{pgfplots}
\pgfplotsset{compat=1.17}
\usepackage[colorlinks=true,linkcolor=black,citecolor=blue!50!black,urlcolor=blue!50!black]{hyperref}

\definecolor{alloyink}{HTML}{1F2A37}
\definecolor{alloyaccent}{HTML}{2F6FDE}
\definecolor{alloysoft}{HTML}{EEF2F7}
\newcommand{\code}[1]{\texttt{#1}}

\title{\Large\textbf{Alloy: A Multi-Material Physics Engine for Robot Learning}}
\author{Pranay Gupta\thanks{Correspondence: \href{mailto:pranay@argonrobotics.ai}{pranay@argonrobotics.ai}.
Code: \url{https://github.com/argonrobotics-ai/alloy}. Docs: \url{https://alloy.argonrobotics.ai/docs/}.} \quad Deepanshu Rohilla\\[2pt]
\normalsize Argon Robotics}
\date{November 2026}

\begin{document}
\maketitle

\begin{abstract}
Robots handle much more than rigid objects: towels, cables, water, sand. The simulators used to train robots at
scale are strongest on rigid bodies, and adding another material usually means a second engine and hand-written
glue between the two. We present Alloy, an open-source physics engine for robot learning in which rigid robots,
cloth, cables, liquids, granular material and soft bodies share one scene. Each object is described once; Alloy
assigns it to a solver suited to its material, couples the solvers inside every control step, and runs the scene
in many parallel worlds on one GPU, or on a CPU. Every world carries
a health grade, to help tell a failure of the physics from a failure of the policy. Alloy is built on Newton, MuJoCo Warp and Warp, and it is
written to be driven by coding agents: its documentation ships inside the package, every error names its fix, and
the engine is exposed as tools over the Model Context Protocol. In a cold test, agents that had never seen Alloy installed a pre-release from the package index and trained policies for walking, cable handling and pouring on a single GPU.
We also report how throughput grows with the number of worlds, depth-camera throughput, agreement with closed-form
references, and a reliability test over a thousand worlds.
\end{abstract}

\begin{figure}[h]
\centering
\begin{subfigure}{0.49\linewidth}\includegraphics[width=\linewidth]{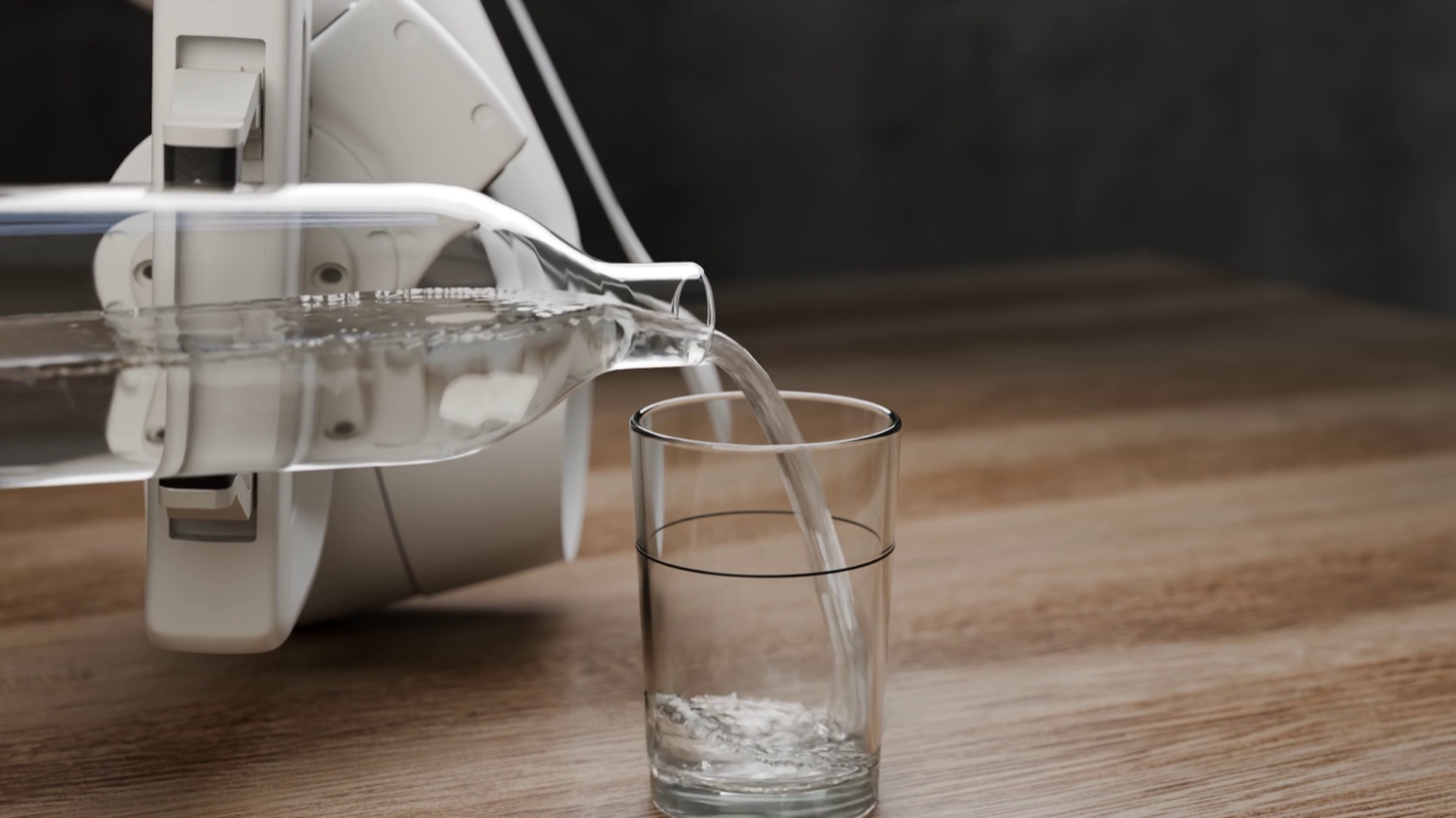}\caption{Pouring water}\end{subfigure}\hfill
\begin{subfigure}{0.49\linewidth}\includegraphics[width=\linewidth]{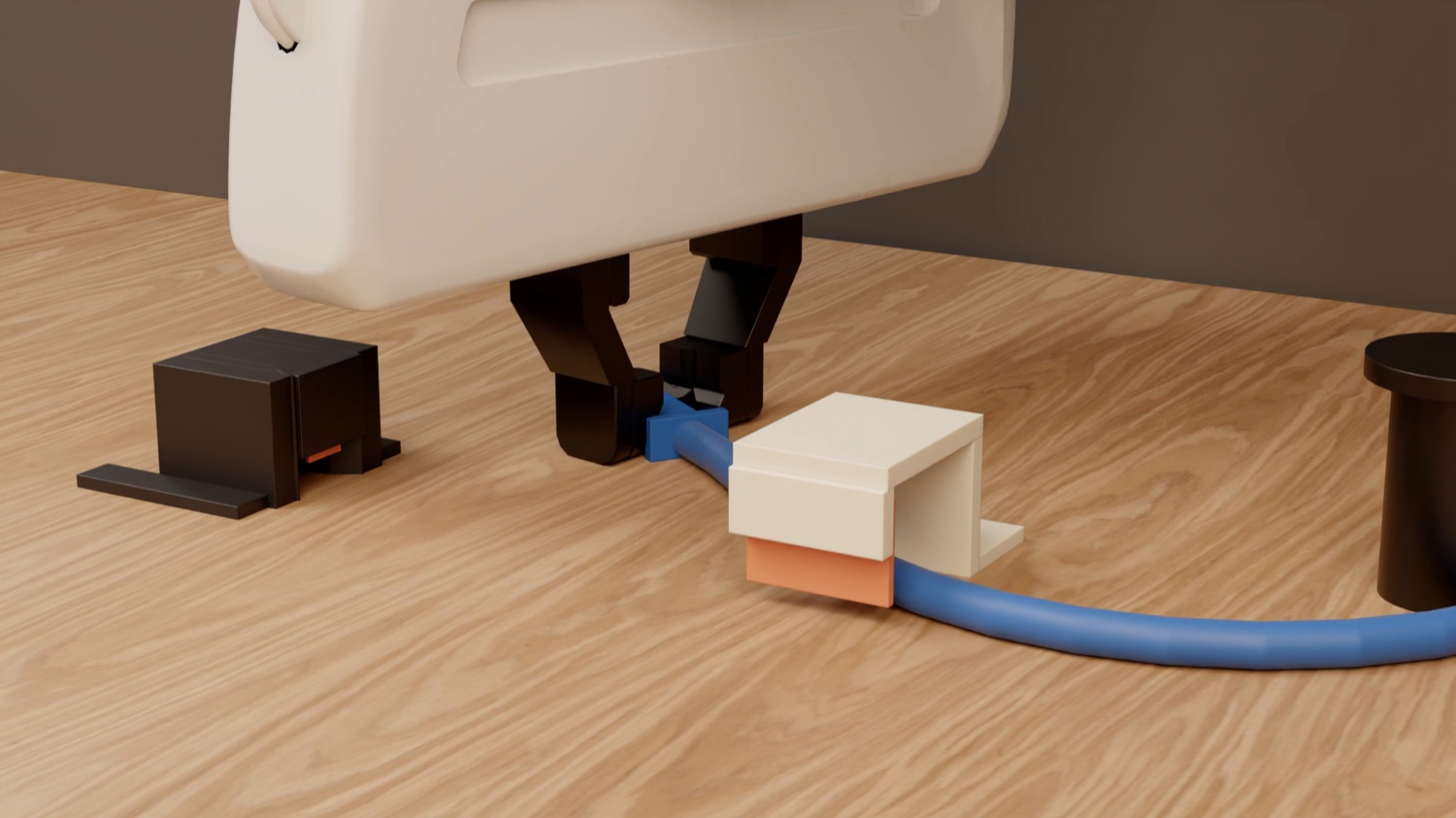}\caption{Routing a cable}\end{subfigure}\\[4pt]
\begin{subfigure}{0.49\linewidth}\includegraphics[width=\linewidth]{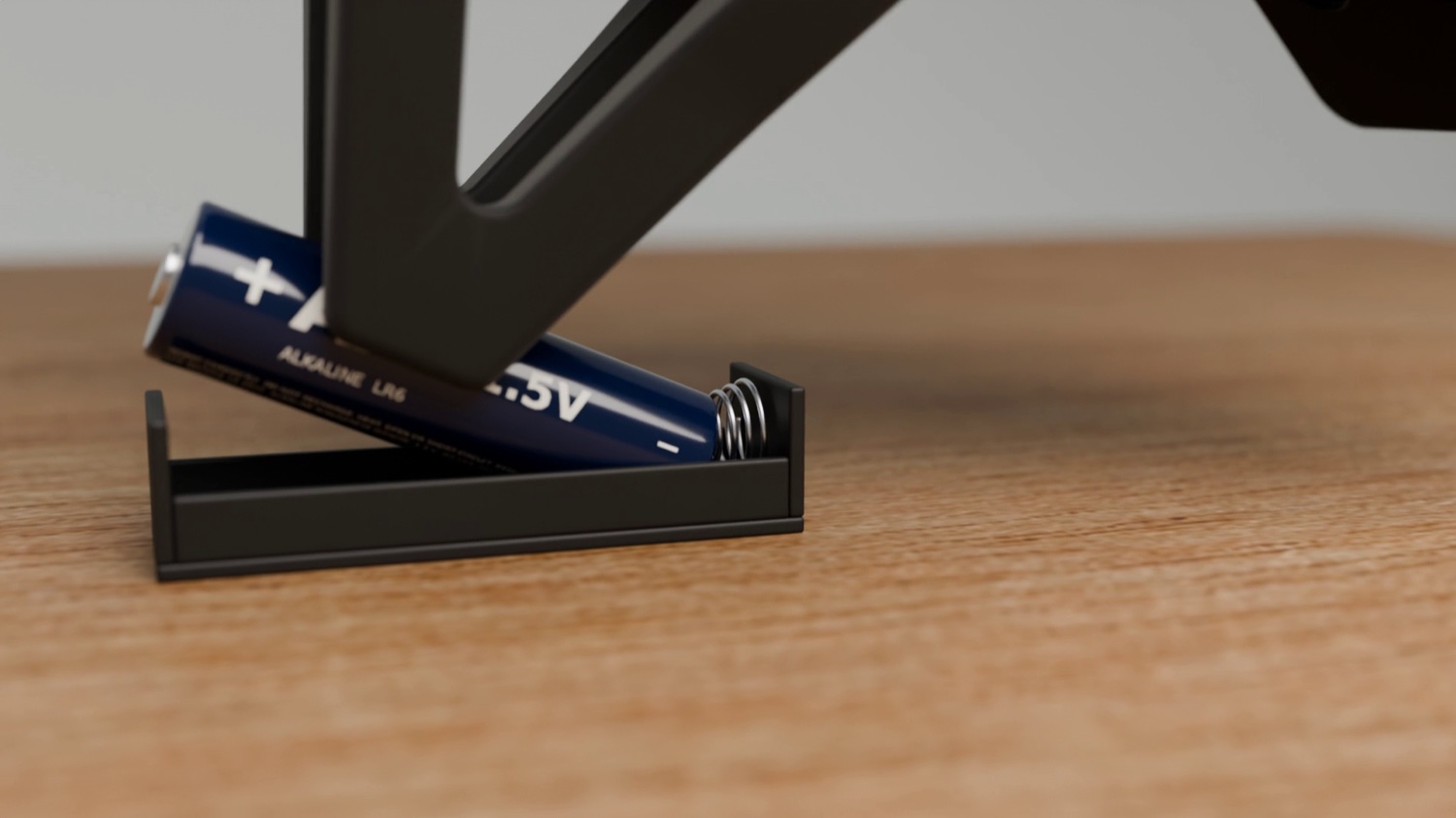}\caption{Inserting a cell into a sprung holder}\end{subfigure}\hfill
\begin{subfigure}{0.49\linewidth}\includegraphics[width=\linewidth]{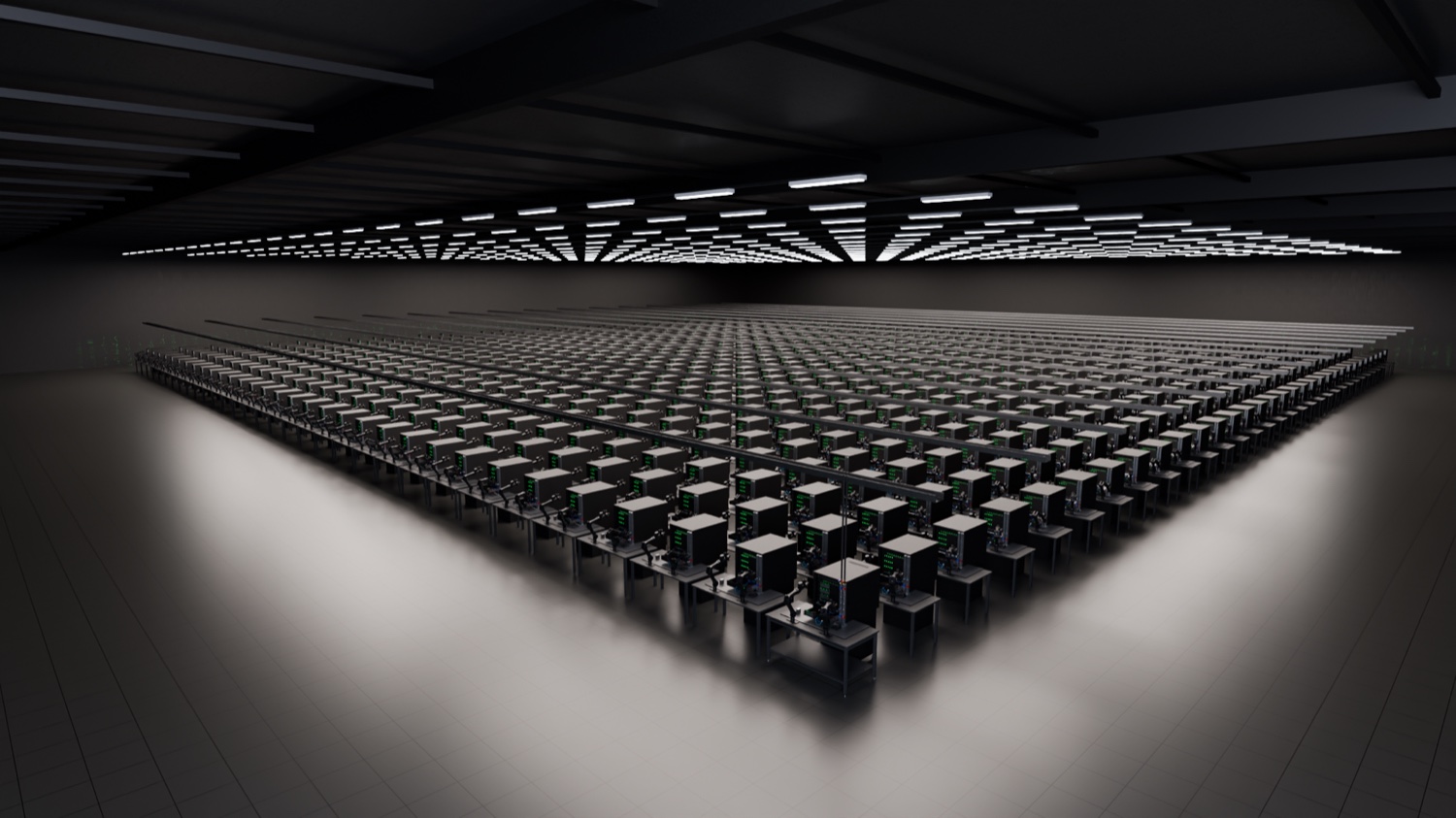}\caption{1,024 parallel worlds of one task}\end{subfigure}
\caption{Scenes simulated in Alloy. (a) A Franka arm pours water from a bottle into a glass; the water is simulated
with Alloy's liquid solver, coupled to the robot's rigid-body contact. (b) A Franka arm routes a cable through a clip
toward a socket; the cable is an articulated chain in the rigid-body solver. (c) A gripper pushes an AA cell into a
sprung battery holder, simulated with rigid-body contact. (d) 1,024 parallel worlds of one robot task in a single frame. Panels (a), (c) and (d) are rendered with Blender
Cycles, (b) with Alloy's path-traced camera.}
\label{fig:teaser}
\end{figure}

\section{Introduction}

Much of what a useful robot must do involves things that are not rigid. A kitchen robot pours and wipes. A factory
robot routes cables and seats connectors. A laundry robot folds. Learning these skills in simulation needs two
things at once. The first is physics for each material: cloth that drapes, liquid that flows, cables that bend. The
second is scale: reinforcement learning tries a task millions of times, so the simulation has to run in thousands
of copies at once and stay correct in every one.

Today these two needs pull apart. Engines built for robot learning at scale are at their best with rigid bodies.
Solvers for cloth, liquids or granular material often come as separate programs with their own scene
descriptions. A team that wants a robot arm to pour water often writes its own bridge between a rigid-body
engine and a liquid solver, chooses time steps by hand, and builds its own checks for worlds that have quietly
gone wrong.

There is also a new kind of user. More and more simulation code is written by coding agents, which read
documentation, write a scene, run it and read the errors. An engine that agents can use well needs short
documentation that travels with the package, errors that say what to change, and a way to tell a broken simulation
from a bad policy.

Alloy is our answer to these three problems. Its contributions are:
\begin{itemize}\itemsep2pt
\item \textbf{One scene for many materials.} Rigid robots, cloth, cables, liquids, granular material and soft
bodies are added through one Python interface. A \emph{router} assigns each object to a solver suited to it and
records why (Section~\ref{sec:routing}).
\item \textbf{Coupling inside every control step.} Solvers exchange poses and contact forces through proxy bodies,
with step sizes chosen from the stiffness of the interface and refined per world when needed
(Sections~\ref{sec:coupling} and~\ref{sec:clocks}).
\item \textbf{A health grade for every world.} Each parallel world is graded \code{OK}, \code{DEGRADED} or
\code{INVALID}; an invalid world is frozen while the others continue (Section~\ref{sec:health}).
\item \textbf{An interface made for coding agents.} Documentation inside the package, errors that carry their fix,
a command-line tool that speaks JSON, and a tool server for the Model Context Protocol
(Section~\ref{sec:agents}).
\item \textbf{Evidence from cold users.} Agents that had never seen Alloy installed it and trained policies on a
single GPU; we report those results with their protocol, together with scaling, camera, accuracy and reliability
measurements (Section~\ref{sec:experiments}).
\end{itemize}
Alloy is open source under the Apache-2.0 licence and installs with \code{pip install alloy-engine}.

\section{Background and Related Work}

\paragraph{Rigid-body engines for robot learning.} MuJoCo~\cite{todorov2012mujoco} is the reference engine for
articulated robots with contact. MuJoCo Warp~\cite{mujocowarp} runs the MuJoCo model on a GPU using
Warp~\cite{macklin2022warp}, a framework that compiles Python functions into GPU kernels, and MuJoCo
Playground~\cite{zakka2025playground} builds robot-learning environments on top of it. Orbit~\cite{mittal2023orbit} (now Isaac Lab) and ManiSkill3~\cite{tao2024maniskill3} provide large sets of GPU-parallel environments for manipulation and
locomotion. Massively parallel simulation made it possible to train legged locomotion in minutes~\cite{rudin2022walk}.

\paragraph{Solvers for other materials.} Vertex Block Descent (VBD)~\cite{chen2024vbd} simulates cloth and soft
bodies by solving for one vertex at a time and is stable with large steps. Smoothed-particle hydrodynamics (SPH)
represents a liquid as particles; its divergence-free variant~\cite{bender2015dfsph} keeps the liquid's volume
constant. The material point method (MPM), brought to graphics for snow~\cite{stomakhin2013mpm}, handles granular material
and pastes.
Newton~\cite{newton} gathers several such solvers in one GPU framework built on Warp, and
Genesis~\cite{genesis2024} combines multiple solvers for robotics in one platform.

\paragraph{Where Alloy sits.} Alloy does not replace these solvers; it is built on them. Rigid bodies and robots
step in MuJoCo Warp on a GPU and in MuJoCo on a CPU; cloth, soft bodies and elastic rods step in Newton's VBD; sand
and pastes step in Newton's implicit MPM; liquids step in Alloy's own SPH solver. What Alloy adds is the layer
around them: one scene description, automatic routing, coupling with automatic step selection, health grading per
world, cameras for every world, and an interface designed for agents. A robot model written for MuJoCo (an MJCF
file) loads in Alloy as it is.

\section{System}

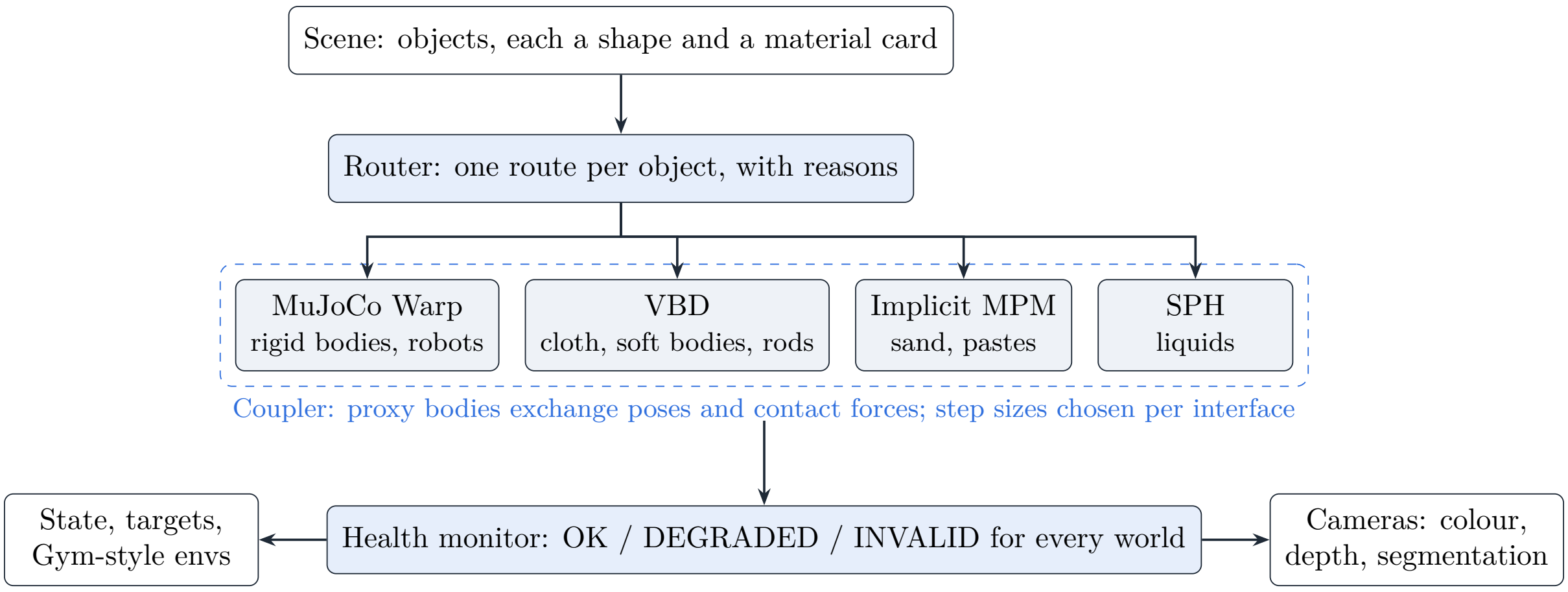
\begin{figure}[t]
\centering
\resizebox{\linewidth}{!}{%
\begin{tikzpicture}[
  box/.style={draw=alloyink, rounded corners=3pt, minimum height=8mm, inner sep=5pt, font=\small, align=center},
  solver/.style={box, fill=alloysoft, minimum width=23mm},
  arr/.style={-{Stealth[length=2mm]}, thick, alloyink}]
\node[box, minimum width=50mm] (scene) {Scene: objects, each a shape and a material card};
\node[box, below=7mm of scene, minimum width=50mm, fill=alloyaccent!12] (router) {Router: one route per object, with reasons};
\node[solver, below left=9mm and 30mm of router.south, anchor=north] (mj) {MuJoCo Warp\\ \footnotesize rigid bodies, robots};
\node[solver, right=3mm of mj] (vbd) {VBD\\ \footnotesize cloth, soft bodies, rods};
\node[solver, right=3mm of vbd] (mpm) {Implicit MPM\\ \footnotesize sand, pastes};
\node[solver, right=3mm of mpm] (sph) {SPH\\ \footnotesize liquids};
\node[draw=alloyaccent, dashed, rounded corners=4pt, inner sep=5pt, fit=(mj)(vbd)(mpm)(sph),
      label={[font=\footnotesize, text=alloyaccent]below:Coupler: proxy bodies exchange poses and contact forces; step sizes chosen per interface}] (coup) {};
\node[box, below=14mm of coup, minimum width=50mm, fill=alloyaccent!12] (health) {Health monitor: OK / DEGRADED / INVALID for every world};
\node[box, right=8mm of health, minimum width=30mm] (cam) {Cameras: colour,\\ depth, segmentation};
\node[box, left=8mm of health, minimum width=30mm] (api) {State, targets,\\ Gym-style envs};
\draw[arr] (scene) -- (router);
\draw[arr] (router.south) -- ++(0,-4mm) -| (mj.north);
\draw[arr] (router.south) -- ++(0,-4mm) -| (vbd.north);
\draw[arr] (router.south) -- ++(0,-4mm) -| (mpm.north);
\draw[arr] (router.south) -- ++(0,-4mm) -| (sph.north);
\draw[arr] (coup.south |- health.north) ++(0,10mm) -- (health.north);
\draw[arr] (health) -- (cam);
\draw[arr] (health) -- (api);
\end{tikzpicture}}
\caption{How a scene becomes a simulation. Each object is routed to one solver. When a scene uses more than one
solver, a coupler joins them inside every control step. A health monitor grades every parallel world after each
exchange.}
\label{fig:arch}
\end{figure}

This section describes the engine as of version 0.1.2. Figure~\ref{fig:arch} gives the overview.

\subsection{Scenes and cards}

A scene is a list of objects. Each object is a shape together with a \emph{material card}: a small file that holds
the material's measured values (density, stiffness, friction and so on) and their sources. The user never names a
solver. Listing~\ref{lst:scene} puts a robot arm and a tub of water in one scene; the arm is written in MuJoCo's
MJCF format.

\begin{lstlisting}[float=t,caption={A robot arm and water in one scene. The printed output shows the solver chosen for each
object, the number of water particles and the joint angle, and the health of the world.},label={lst:scene}]
import alloy

ARM = """<mujoco><compiler angle="radian"/><worldbody><body name="link1" pos="0 0 0.1">
  <joint name="shoulder" type="hinge" axis="0 1 0" range="-2 2" damping="2"/>
  <geom type="capsule" fromto="0 0 0 0.3 0 0" size="0.02" mass="1"/>
  </body></worldbody>
  <actuator><position joint="shoulder" kp="200" kv="10"/></actuator></mujoco>"""

scene = alloy.Scene(dt=0.01)
scene.add_ground()
scene.add_robot("arm", mjcf=ARM, pos=(0, 0, 0.1))
scene.add_open_box("tub", inner=(0.08, 0.08, 0.05), wall=0.01, pos=(0.3, 0, 0))
scene.add_liquid("water", size=(0.064, 0.064, 0.024), pos=(0.3, 0, 0.024))
sim = alloy.build(scene, device="cpu")
print({name: sim.routes()[name]["solver"] for name in ("arm", "water")})
sim.run(0.6, targets={"shoulder": (0.0, 0.3)})
print(len(sim.state(["water"], world=0)["water.points"]),
      sim.state(world=0)["arm/shoulder.q"].round(2))
print(sim.check()["ok"], sim.health()["worlds"][0]["severity"])

# {'arm': 'mujoco', 'water': 'sph'}
# 1536 0.28
# True OK
\end{lstlisting}

\subsection{Routing}\label{sec:routing}

When the scene is built, the router gives every object one \emph{route}. It decides from physical checks on the
object, in a fixed order. Static objects need no solver. Next comes phase: liquids, pastes and granular material
are recognised from the card. Next comes shape class: rod, sheet or volume. Last comes rigidity: the router
estimates how far the object would be squeezed under the grip force named on its card, compares that with the
card's tolerance, and checks how much the whole body would bend. An object that passes is treated as rigid; one
that fails is deformable. A user can override the route for any object.

Routes map to solvers as in Table~\ref{tab:routes}. Every decision records the numbers it used and its margin to
the nearest threshold, and \code{sim.routes()} returns the route, the solver and the reasons. A user, or an agent,
can therefore see why a towel became cloth and why a steel ruler did not.

\begin{table}[H]
\centering\small
\begin{tabular}{@{}l>{\raggedright\arraybackslash}p{0.56\linewidth}@{}}
\toprule
Solver & Objects routed to it \\
\midrule
MuJoCo Warp (GPU), MuJoCo (CPU) & rigid bodies, robots, unclamped chains (such as cables), creased panels \\
VBD (Newton) & cloth, soft volumes, elastic rods \\
Style3D (Newton) & anisotropic cloth when it is the only dynamic object \\
Film (Alloy, on VBD) & paper, polymer film and metal sheets \\
Implicit MPM (Newton) & sand, pastes, viscous or yielding fluids \\
SPH (Alloy) & liquids \\
\bottomrule
\end{tabular}
\caption{Solvers and what the router sends to them.}
\label{tab:routes}
\end{table}

\subsection{Coupling}\label{sec:coupling}

If every dynamic object in a scene uses the same solver, that solver steps alone and there is nothing to couple.
When two or more solvers are present, they are joined in pairs, each with a \emph{source} and a \emph{destination};
by default the rigid-body solver is the source.

The mechanism is a \emph{proxy body}, using Newton's proxy coupler with Alloy's scheduler on top. The bodies of the source appear inside the destination as proxies that carry
the source's poses and velocities. The destination solves its own contacts against these proxies. The contact
forces it applied to them are then collected and fed back to the source. In this way a towel pushes on a robot's gripper and the gripper pushes on the towel, while each solver keeps its own
contact model. A pair can also be one-way: the destination follows the source and no force is fed back. This is
used for a liquid whose mass is small beside the rigid bodies that drive it.

One such round is an \emph{exchange}. Inside an exchange every solver steps once, in an order fixed by the graph of
pairs. The default mode runs two passes. A staggered mode repeats passes while some world's \emph{interface
residual}, the disagreement between the two sides of a pair, is above tolerance, up to a cap. In the staggered mode the fed-back forces are relaxed per world with Aitken's
method~\cite{kuttler2008aitken}, which adapts the relaxation factor from one pass to the next. A pair is
\emph{strong} when the destination adds more than half of the source's mass. Strong pairs into a liquid run the
staggered mode; other strong pairs, and most scenes with three or more solvers, use Newton's ADMM
coupler~\cite{boyd2011admm}. Residuals are computed per world with
order-independent operations, and a world that has converged is frozen, so its result does not depend on which
other worlds share the batch.

\subsection{Choosing step sizes}\label{sec:clocks}

A control period of length $\Delta t$ (for example 10\,ms) is split into $2^{j}$ exchanges. The exchange interval
$h = \Delta t / 2^{j}$ is the largest one that satisfies
\begin{equation}
h \;\le\; s \,\, \omega_{\max} \sqrt{\frac{m_\mathrm{eff}}{k_c}},
\label{eq:clock}
\end{equation}
where $m_\mathrm{eff}$ is the effective mass at the interface between the two solvers, $k_c$ is the contact
stiffness there, $\omega_{\max}$ is a stability constant (the largest allowed product of the interface's natural frequency and the
step), and $s<1$ is a safety factor. Further caps apply, such as a maximum exchange step.
In words: a stiff contact between light objects oscillates quickly and needs frequent exchanges; a soft contact
between heavy objects does not. Inside one exchange each solver then takes as many sub-steps as its own stability
rule asks for; the liquid solver, for instance, uses its own limit on how far a particle may move per step.

The clock moves to a finer step at once and returns to a coarser one only after ten quiet periods. A watchdog
books the work done across every two-way interface. If that work is out of proportion to the world's energy, or
the two sides of an interface drift apart, the world \emph{escalates}: it returns to the start of the control
period and runs that period again with a finer step. On a first
escalation only that world is run again, and every other world keeps its first result bit for bit; repeated
escalations make the step finer for the whole scene.

\subsection{A health grade for every world}\label{sec:health}

With thousands of worlds, some will meet conditions the solvers handle poorly, and a learning algorithm will
happily exploit a broken world. After every exchange Alloy's health monitor checks each world for non-finite
values, penetration, excessive speed, jumps in energy, and exceeded joint or buffer limits. Each world is graded
\code{OK}, \code{DEGRADED} or \code{INVALID}. An invalid world is returned to a saved good state (by default the start of its control period) and frozen while the
others continue. \code{sim.check()} returns one value to read after a run, and
\code{sim.health()} returns one row per world with the flags raised and, for each, the change that would fix it.
Training code can use these grades to separate ``the physics failed'' from ``the policy failed''.

\subsection{Many worlds, cameras and robots}

A scene is replicated into many worlds that step together on one GPU; the same script runs on a CPU. On a GPU, chunks of up to 64 exchanges, with their sub-steps and coupler passes, are captured as graphs and
replayed without returning to the host in between. In a scene with robots, rigid bodies step at a fixed 1.25\,ms by
default at a 10\,ms control period; a scene of only free rigid bodies takes the largest stable step. Tight fits such as
pegs, threads and connectors can be declared and receive extra contact handling. A camera can be attached to every
world and returns colour, depth and segmentation images. Environments follow the shape of Gymnasium's vector interface~\cite{towers2024gymnasium}: every world is one
environment.

The package ships nine robots (two Franka arms, a UR10e with a two-finger gripper, two bimanual Franka setups, the
Unitree G1 and H1-2 humanoids, the Unitree Go2 quadruped and the LEAP hand; most models come from MuJoCo
Menagerie~\cite{menagerie}), nine ready-made environments, legged-locomotion environments, 90 material cards and
118 object cards. A further library of 5,423 object cards can be downloaded.

\subsection{An interface for coding agents}\label{sec:agents}

Alloy is written to be driven by an agent as well as by a person.
\begin{itemize}\itemsep2pt
\item \textbf{Documentation travels with the package.} \code{python -m alloy.docs} prints the location of the
guides, the runnable examples and an agent guide that were installed with the package, so an agent reads the
documentation for exactly the version it is running.
\item \textbf{Errors carry their fix.} Every error has a message and a \code{fix} field that says what to change;
a misspelt name lists the valid ones.
\item \textbf{A command-line tool that speaks JSON.} Every reply is one JSON object with either a result or an
error, its fix and the field at fault.
\item \textbf{A tool server.} \code{alloy mcp} serves the engine over the Model Context
Protocol~\cite{mcp}: an agent can list the library, build a scene, step it, read state, set targets, check health
and render an image as tool calls.
\item \textbf{Helpers for building scenes.} \code{alloy.compose} places library objects from a sentence and checks
the result with a short simulation; \code{alloy.create} turns a mesh, a photo or a description into a checked
object card.
\end{itemize}

\section{Experiments}\label{sec:experiments}

\subsection{Can a newcomer train with it?}\label{sec:cold}

An engine for agents should be judged in the hands of an agent that has never seen it. We ran a cold test.
Eight independent sessions of a coding agent (Claude Opus 5.5) each started on a fresh virtual machine with one
NVIDIA L4 GPU. Each installed Alloy
from the Python package index and was allowed to read only what the package ships. Each wrote its own training and
evaluation code. Success was computed from raw simulator state by the tester's own script, on episodes whose random
seeds differ from the training seed; we call these held-out episodes. 

Table~\ref{tab:train} lists the results. All of them were measured on \code{alloy-engine} 0.1.0rc2, the pre-release that preceded the version described
in Section~3, each from one training run. Intervals are Wilson 95\,\% intervals.

\begin{table}[!t]
\centering\small
\begin{tabular}{@{}>{\raggedright\arraybackslash}p{0.30\linewidth}>{\raggedright\arraybackslash}p{0.22\linewidth}>{\raggedright\arraybackslash}p{0.13\linewidth}>{\raggedright\arraybackslash}p{0.26\linewidth}@{}}
\toprule
Task & Training & Training time & Held-out success \\
\midrule
\textbf{Walking}: Unitree Go2, shipped environment, flat ground & PPO, shipped trainer, 4,096 worlds
  & 4.1\,min\newline 15.5\,min & 244/256, 95\,\% [92, 97]\newline 497/512, 97\,\% [95, 98] \\\addlinespace
\textbf{Push recovery}: the same policy, pushed with 150\,N for 0.1\,s & as above & 15.5\,min & 252/256 and 248/256, 97--98\,\% \\\addlinespace
\textbf{Cable lift}: task written by the tester & PPO, shipped trainer, 1,024 worlds
  & about 15\,min\newline about 30\,min & 242/256, 94.5\,\% [91, 97]\newline 245/256, 95.7\,\% [93, 98] \\\addlinespace
\textbf{Pouring}: shipped environment & PPO, 512 worlds & 38.5\,min & 118/128, 92\,\% [86, 96] \\\addlinespace
\textbf{Cup pouring}: task written by the agent, on its second reward version & PPO, 256 worlds & about 70\,min & 256/256, 100\,\% [98.5, 100] \\\addlinespace
\textbf{Pick-and-place by imitation}: shipped environment, gantry gripper, state observations
  & behaviour cloning, 1,000 scripted demonstrations & 66\,s & 256/256, 100\,\% [98.5, 100] \\
\bottomrule
\end{tabular}
\caption{Policies trained by cold agents on one NVIDIA L4 with \code{alloy-engine} 0.1.0rc2; one training run per
row; Wilson 95\,\% intervals in brackets. The two walking rows are the same run after 100 and 400 iterations, the
second evaluated on two seeds; its times exclude the one-time compilation of GPU kernels. No evaluated episode in
the walking or cable rows was graded \code{INVALID}.}
\label{tab:train}
\end{table}

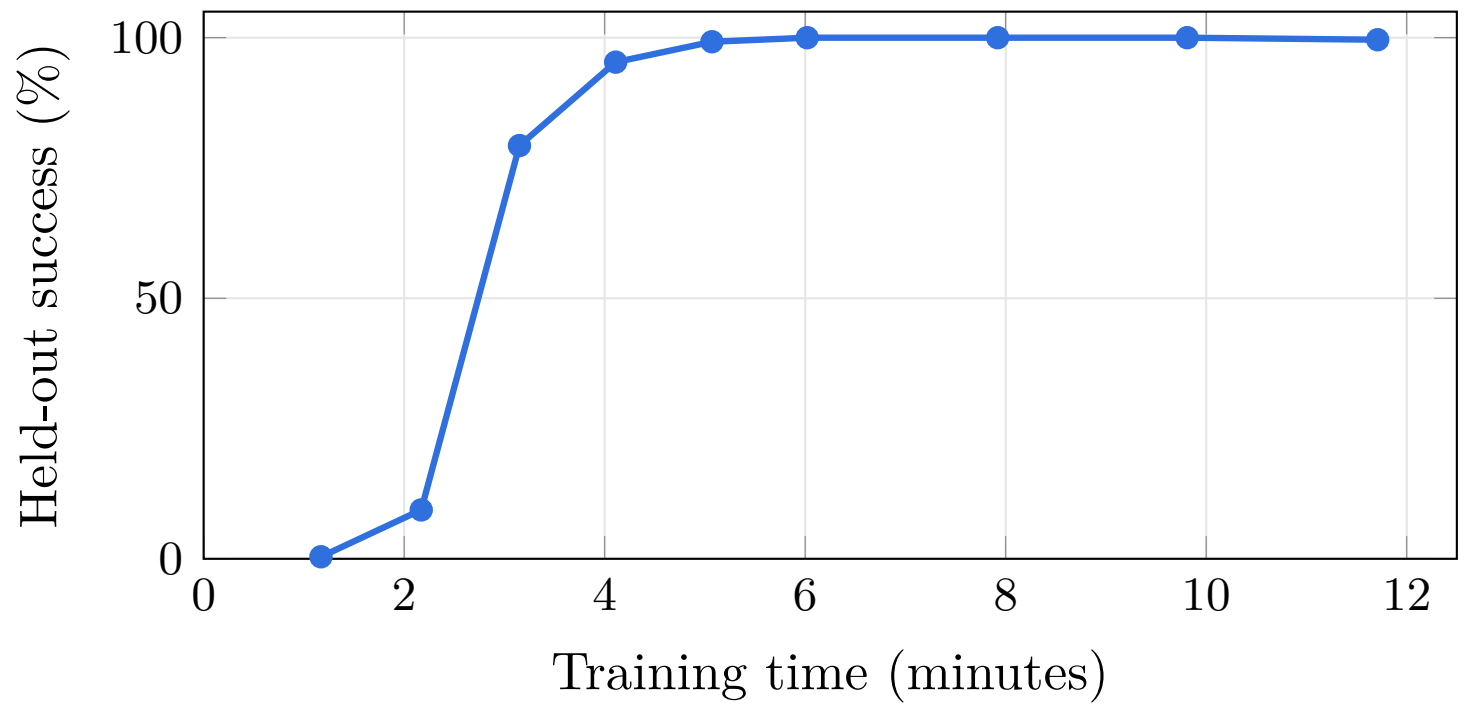
\begin{figure}[H]
\centering
\begin{tikzpicture}
\begin{axis}[width=0.62\linewidth, height=5.2cm, xlabel={Training time (minutes)}, ylabel={Held-out success (\%)},
  ymin=0, ymax=105, xmin=0, xmax=12.5, grid=major, grid style={gray!20}, tick label style={font=\footnotesize},
  label style={font=\small}]
\addplot[alloyaccent, very thick, mark=*, mark size=1.6pt] coordinates
 {(1.17,0.39) (2.17,9.38) (3.15,79.3) (4.11,95.31) (5.07,99.22) (6.02,100) (7.92,100) (9.81,100) (11.71,99.61)};
\end{axis}
\end{tikzpicture}
\caption{Go2 walking on one L4 with 4,096 worlds: held-out success against training time. Each point is 256
episodes on one evaluation seed. The one-time compilation of GPU kernels (202\,s on first use) is not counted.}
\label{fig:go2}
\end{figure}

\paragraph{Walking.} The task is the shipped velocity-tracking environment for the Unitree Go2 on flat ground,
trained with PPO~\cite{schulman2017ppo} through the shipped trainer, which uses the \code{rsl\_rl}
library~\cite{rudin2022walk}. An episode succeeds if the robot does not fall and the world is not graded \code{INVALID} in 20\,s, and its mean
velocity stays
within 0.25\,m/s, and its turning rate within 0.25\,rad/s, of the command over the last 15\,s. Figure~\ref{fig:go2}
shows held-out success against training time. The policy after 400 iterations was then pushed with 150\,N for 0.1\,s in a random direction while walking at
0.5\,m/s, a harder push than any it met in training (at most 113\,N). It was still standing at the end of the
episode in 252 and 248 of 256 episodes on two seeds.

\paragraph{Cable lift.} A tester wrote this task: lift a cable off a table. An episode succeeds if every point of the cable's centre line is above 3\,cm at the last step. The shipped
one-command trainer reached 94.5\,\% after about 15 minutes and 95.7\,\% after about 30.

\paragraph{Pouring.} In the shipped pouring environment an episode succeeds if at least 75\,\% of the water ends in
the bowl. The checkpoint saved at 38.5 minutes of training succeeded in 118 of 128 held-out episodes. Another
agent wrote its own cup-pouring task; on its second reward version, after about 70 minutes of training, the policy
passed the same 75\,\% bar in 256 of 256 held-out episodes.

\paragraph{Imitation.} On the shipped pick-and-place environment with a gantry gripper and state observations, a
tester collected 1,000 scripted demonstrations from 256 worlds and trained a three-layer network by behaviour
cloning in 66 seconds. The object ended in the bin in 256 of 256 held-out episodes.

\subsection{Throughput against the number of worlds}

\begin{figure}[H]
\centering
\begin{tikzpicture}
\begin{loglogaxis}[width=0.66\linewidth, height=5.6cm, xlabel={Parallel worlds},
  ylabel={Simulated seconds per wall second}, log basis x=2, xtick={1,4,64,512}, xticklabels={1,4,64,512},
  grid=major, grid style={gray!20}, tick label style={font=\footnotesize}, label style={font=\small},
  legend style={font=\footnotesize, at={(1.03,0.5)}, anchor=west, draw=none}, legend cell align=left]
\addplot[alloyaccent, very thick, mark=*] coordinates {(1,0.83) (4,3.3) (64,48) (512,313)};
\addlegendentry{Three objects in a bin}
\addplot[alloyink, very thick, mark=square*] coordinates {(1,1.9) (4,6.9) (64,74)};
\addlegendentry{Franka Panda scene}
\addplot[orange!80!black, very thick, mark=triangle*] coordinates {(1,0.53) (4,2.0) (64,7.1)};
\addlegendentry{200 ml pour}
\addplot[green!45!black, very thick, mark=diamond*] coordinates {(1,0.20) (4,0.73) (64,4.1)};
\addlegendentry{Cotton sheet on a table}
\end{loglogaxis}
\end{tikzpicture}
\caption{Throughput against world count on one NVIDIA L4: physics only, 10\,ms control period, one process, health
check passed at every point. Rates are warm: control periods in which GPU graphs were first captured are replaced
by the median of later periods. Four of the nine scenes of a calibration set are shown, measured on a development
build before the first release; one run per point.}
\label{fig:scale}
\end{figure}
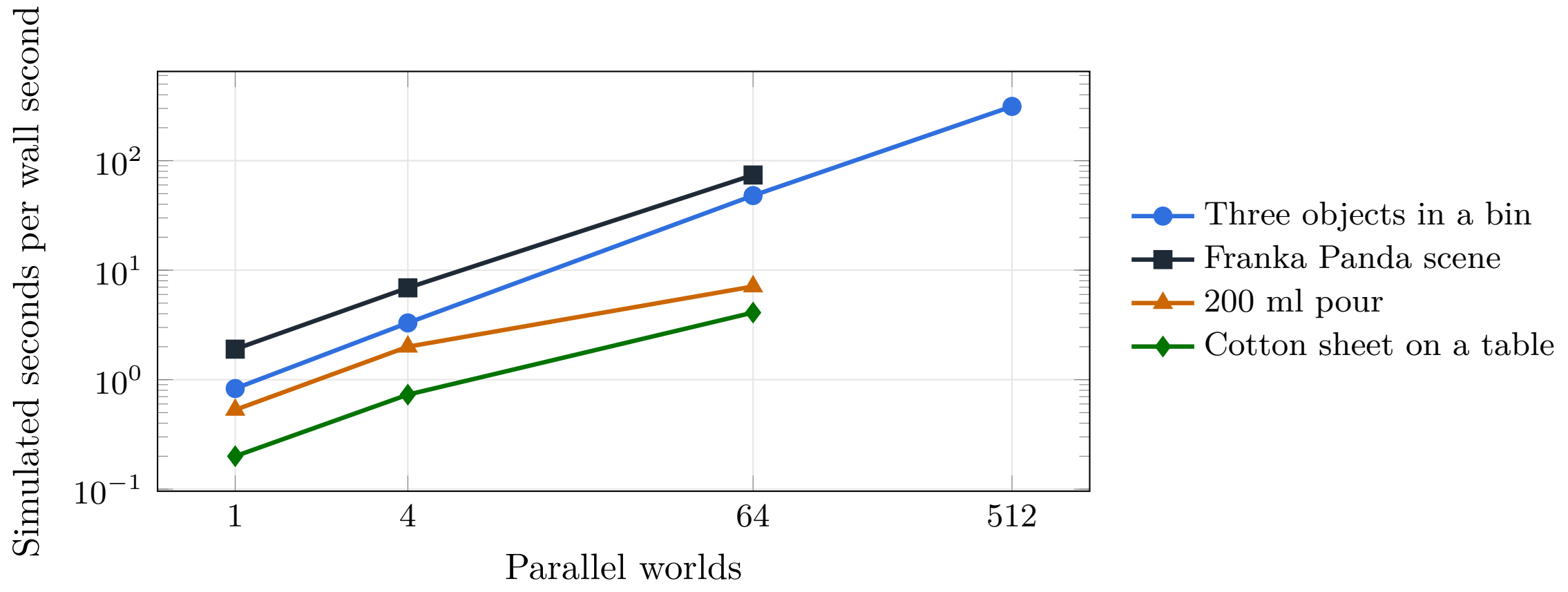

Figure~\ref{fig:scale} shows how throughput grows when one scene is replicated into more worlds on one NVIDIA L4.
Throughput is the number of simulated seconds produced per second of wall time, summed over the worlds. From 1 to
64 worlds it grows about 39 times for a Franka Panda scene, about 20 times for a cotton sheet on a table and about
13 times for a 200\,ml pour; for three objects dropped in a bin it grows about 377 times from 1 to 512 worlds.

\subsection{Depth cameras for every world}

Policies that learn from images need a camera in every world. Table~\ref{tab:depth} gives the rate at which Alloy
renders depth images for 1,024 worlds of a Franka scene on one L4, with no physics between renders. A second run
on a different L4 machine is shown beside the first. Adding segmentation to depth changed the cost by less than
1\,\%.

\begin{table}[H]
\centering\small
\begin{tabular}{@{}lrrr@{}}
\toprule
 & $64\times64$ & $128\times128$ & $256\times256$ \\
\midrule
Depth images per second, run 1 & 70,786 & 16,111 & 4,133 \\
Depth images per second, run 2 (another machine) & 62,785 & 14,372 & 4,106 \\
\bottomrule
\end{tabular}
\caption{Depth-image throughput at 1,024 worlds on one NVIDIA L4 (Franka scene, rendering only, warm), measured
on a development build before the first release.}
\label{tab:depth}
\end{table}

\subsection{Agreement with closed-form references}

Where physics has a closed-form answer, a simulator can be checked against it. Table~\ref{tab:laws} lists ten
selected checks, each with the error Alloy measured against the reference and Alloy's own tolerance for the check, set before
it was run.
They cover rigid-body motion and friction, the bending, twisting, sagging and buckling of rods, the pressure of a
liquid at rest and the stress of a soft solid.

\begin{table}[H]
\centering\small
\begin{tabular}{@{}llrr@{}}
\toprule
Check & Reference & Error & Tolerance \\
\midrule
Pendulum period$^\dagger$ & small-angle period & $8.0\times10^{-6}$ & $1\times10^{-3}$ \\
Momentum & conservation of momentum & $3.9\times10^{-6}$ & $1\times10^{-3}$ \\
Sliding with friction & Coulomb friction law & $2.8\times10^{-3}$ & $1\times10^{-2}$ \\
Rope around a post$^\dagger$ & capstan equation & exact & $2\times10^{-2}$ \\
Rod bending under a tip load & cantilever deflection & $6.5\times10^{-7}$ & $1\times10^{-2}$ \\
Rod twisting & torsion formula & $1.1\times10^{-6}$ & $1\times10^{-2}$ \\
Hanging cable$^\dagger$ & catenary sag & $2.0\times10^{-3}$ & $1\times10^{-2}$ \\
Column under load$^\dagger$ & Euler buckling load & $1.3\times10^{-2}$ & $5\times10^{-2}$ \\
Liquid at rest & hydrostatic force on a wall & $1.1\times10^{-2}$ & $3\times10^{-2}$ \\
Soft solid under strain & neo-Hookean stress & $2.8\times10^{-3}$ & $5\times10^{-2}$ \\
\bottomrule
\end{tabular}
\caption{Ten selected closed-form checks: error against the reference and Alloy's own tolerance, set in advance, both as
defined by each check.
All ten ran on a CPU, on a development build before the first release, one run each. $^\dagger$ Run at the time
step set for that check rather than the default step.}
\label{tab:laws}
\end{table}

\subsection{Reliability over a thousand worlds}

A learning run uses every world, so a rare failure matters. We stacked ten 5\,cm blocks into a tower in each of
1,024 worlds on one L4, with the starting positions shifted by up to 0.5\,mm from world to world, and simulated
5\,s. A world counts as an error if any block moves or sinks by more than 1\,cm or any value becomes non-finite.
At each of Alloy's three rigid-body settings (accurate, default and fast), and in two separate runs, 0 of 1,024
worlds had an error.

\section{Scope}

Alloy is an early, 0.x release, and its interfaces may still change. What does not work yet is listed on one public
page, each item with its measured size and what to do instead:
\url{https://alloy.argonrobotics.ai/docs/keep-in-mind/}.

\section{Conclusion}

Alloy lets a robot-learning scene contain what real tasks contain: robots together with cloth, cables, liquids,
granular material and soft bodies. It chooses a solver for each object, couples the solvers inside every control
step with step sizes it derives itself, grades the health of every world, and presents all of this through an interface designed for a coding agent working from the documentation
that ships with the package. In a cold test, agents installed a pre-release and trained policies for walking, cable
handling and pouring on a single GPU. Alloy is open source under Apache-2.0:
\code{pip install alloy-engine}.

\paragraph{Acknowledgements.} Alloy builds on Newton, MuJoCo, MuJoCo Warp and Warp, and ships robot models from
MuJoCo Menagerie. We thank the authors and maintainers of these projects.

\end{document}